\documentclass{article}
\usepackage{spconf,amsmath,graphicx}
\usepackage[T1]{fontenc}
\usepackage{verbatim}
\usepackage{amssymb}
\usepackage{booktabs}
\usepackage{multirow}
\usepackage{soul}
\usepackage{xcolor}
\usepackage[normalem]{ulem}
\usepackage{xcolor}

\usepackage{enumitem}
\setlist{nosep, leftmargin=14pt}

\title{MAGIC: Modality-Aware Gini Inspection for Clinical Hallucination Detection in LMMs}

\name{\begin{tabular}{@{}c@{}}
Sachin Acharya$^{*1}$ \qquad Niyoj Oli$^{*1,4}$ \qquad Shishir Adhikari$^{2}$ \qquad Aminu Lawal$^{6}$ \\
Yash Raj Shrestha$^{5}$ \qquad Prashnna K Gyawali$^{3}$ \qquad Binod Bhattarai$^{4,6}$
\end{tabular}}
\address{$^{1}$Nepal Applied Mathematics and Informatics Institute for research, Nepal \\
$^{2}$Global College of Technology, Nepal\\
$^{3}$University of West Virginia, USA \\
$^{4}$ Fogsphere (Redev.AI Ltd, UK) \\
$^{5}$University of Lausanne, Switzerland \\
$^{6}$University of Aberdeen, UK \\
}

\begin{document}
%
\maketitle
\def\thefootnote{*}\footnotetext{These authors contributed equally to this work}
\begin{abstract}

Recent advances in medical large multimodal models (LMMs) have brought promising progress to clinical decision support, yet hallucination, the generation of confident but clinically unverifiable outputs, poses a critical threat to the reliability demanded in high-stakes medical settings. Existing hallucination detection methods, predominantly designed for unimodal settings, fail to capture the visual-linguistic dynamics inherent to medical imaging. We formulate the problem of hallucination detection in LMMs as disproportionate distribution of attention weights across the modalities during answer token generation. We quantify modality-specific attention concentration across layers and heads using the Gini coefficient and demonstrate statistically significant redistribution of attention mass between visual and textual tokens in hallucinated responses. We leverage these features to train a logistic regression-based classifier and evaluate it on two medical LMMs across three medical imaging modalities, demonstrating that our method achieves competitive or superior performance against contemporary hallucination detection baselines. Code is available at \textit{https://github.com/bhattarailab/MAGIC}.
\end{abstract}
\section{Introduction}
\label{sec:intro}
Large multimodal models (LMMs) are emerging as
powerful tools for clinical reasoning,
enabling automated interpretation of different diagnostic modalities in response to natural language queries \cite{vlm_use_2}. By jointly modeling visual and textual information, these models enable visual question answering (VQA) \cite{mozaffari2024vision}, automated report generation \cite{gu2024medreport}, and clinical decision support \cite{mozaffari2024vision}. Despite these advanced capabilities, LMMs may hallucinate~\cite{huang2023survey} whereby the model fabricates findings, misses critical abnormalities, or misinterprets visual evidence while producing confident and reasonable outputs. In clinical settings, such behavior poses significant risks to patient safety and undermines trust in AI-assisted healthcare systems. Therefore, the ability to reliably detect hallucinations in LMMs remains a critical bottleneck to their safe and widespread clinical adoption.

Hallucinations in LMMs have been attributed to over-reliance on language priors rather than visual evidence~\cite{ref_vcd}, dataset biases~\cite{when_seeing_overrides_knowing}, and spurious correlations~\cite{ref_mitagating hallucination}, all of which manifest as misalignment in cross-modal attention patterns.
Despite this insight, attention weights remain largely underexplored as a direct signal for hallucination detection in LMMs. 
While prior work has investigated spectral analysis of attention weights~\cite{spectral_analysis} and attention-dynamics-based detection~\cite{look_back_lens} in unimodal settings, 
directly extending these methods to LMMs overlooks the fundamentally distinct roles of visual and textual tokens.

\begin{figure}[t]
    \centering
    \includegraphics[width=\linewidth]{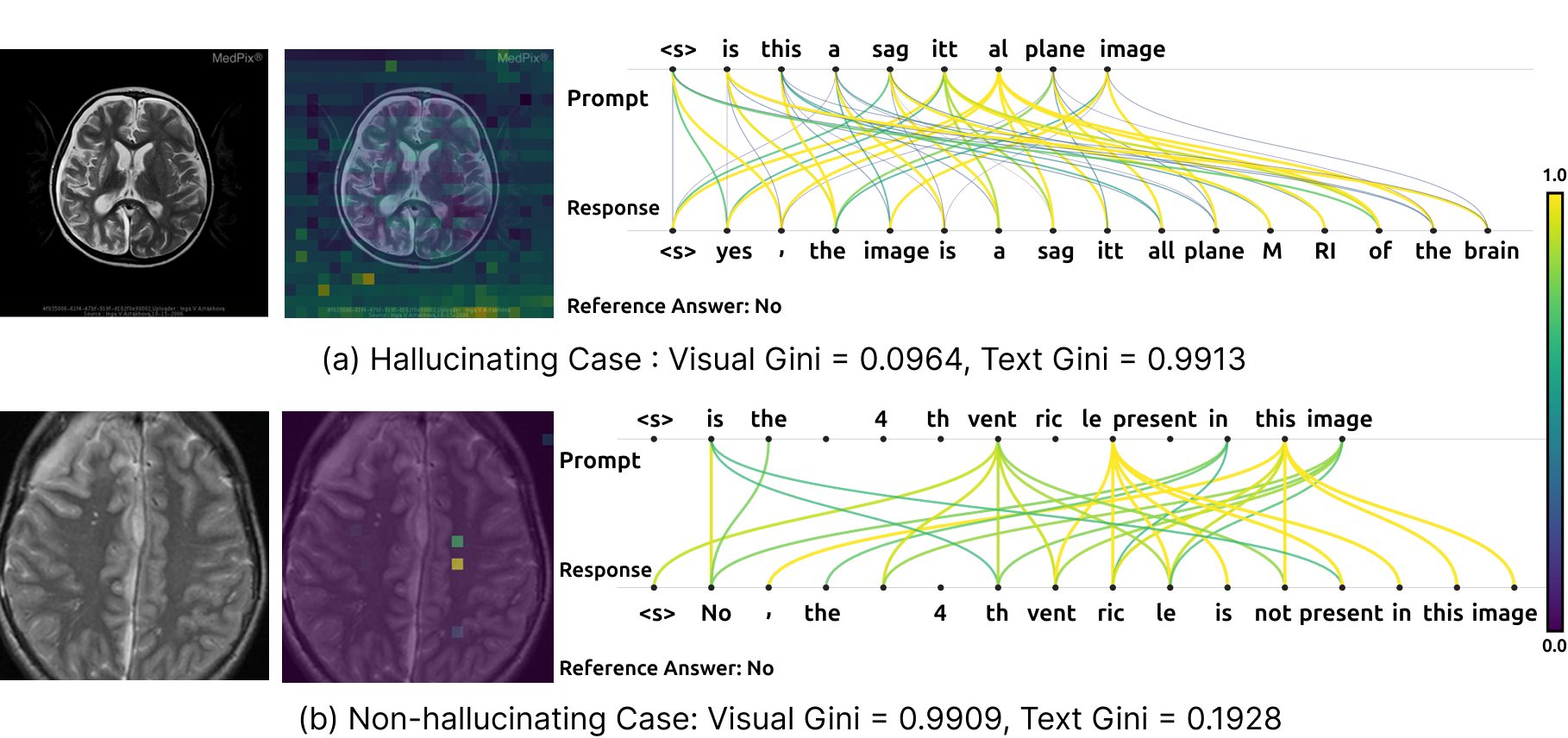}
    \caption{
    Image, visual attention heatmap, and response attention graph (left to right). (a) Hallucinated: diffuse visual attention (low visual Gini) and concentrated text attention (high text Gini). (b) Non-hallucinated: focused visual attention (high visual Gini) and diffuse text attention (low text Gini). 
    }
    \label{fig:gini_visual}
\end{figure}

\begin{table}[t]
\centering
\caption{
Mean Gini coefficients for hallucinated (H) and non-hallucinated (NH) responses, their difference (Diff. = H $-$ NH), and Cohen's $d$ for the five most statistically significant attention heads in each modality. Heads are selected based on p-values comparing the two response groups.
}
\label{tab:gini_attention}
\setlength{\tabcolsep}{3pt}
\resizebox{\linewidth}{!}{
\begin{tabular}{cccccccc}
\toprule
\textbf{Model} & \textbf{Modality} & \textbf{Head ($L,H$)}
& \multicolumn{2}{c}{\textbf{Mean Gini}}
& \textbf{Diff.} & \textbf{Cohen's $d$} \\
\cmidrule(lr){4-5}
& & & \textbf{H} & \textbf{NH} & & \\
\midrule

\multirow{20}{*}{Med-Gemma}
& \multirow{10}{*}{Text}
& 17,2  & 0.8773 & 0.8283 & +0.0490 & +0.719 \\
& & 23,2  & 0.8703 & 0.8304 & +0.0399 & +0.739 \\
& & 17,1  & 0.8215 & 0.7771 & +0.0444 & +0.754 \\
& & 11,3  & 0.8246 & 0.7895 & +0.0351 & +0.724 \\
& & 23,0  & 0.9048 & 0.8685 & +0.0363 & +0.694 \\

\cmidrule(lr){2-7}

& \multirow{10}{*}{Visual}
& 23,5  & 0.9291 & 0.9463 & -0.0171 & -0.704 \\
& & 11,7  & 0.8957 & 0.9237 & -0.0280 & -0.725 \\
& & 17,6  & 0.9311 & 0.9488 & -0.0177 & -0.635 \\
& & 29,2  & 0.8656 & 0.8821 & -0.0165 & -0.632 \\
& & 23,4  & 0.9460 & 0.9628 & -0.0168 & -0.557 \\

\midrule

\multirow{20}{*}{LLaVA-Med-7B}
& \multirow{10}{*}{Text}
& 22,29 & 0.9544 & 0.9494 & +0.0050 & +0.509 \\
& & 16,29 & 0.9667 & 0.9617 & +0.0050 & +0.490 \\
& & 19,10 & 0.9252 & 0.9147 & +0.0105 & +0.492 \\
& & 9,2   & 0.9670 & 0.9645 & +0.0025 & +0.456 \\
& & 19,9  & 0.9490 & 0.9423 & +0.0067 & +0.470 \\

\cmidrule(lr){2-7}

& \multirow{10}{*}{Visual}
& 18,14 & 0.8885 & 0.8775 & +0.0111 & +0.389 \\
& & 27,6  & 0.8979 & 0.9096 & -0.0117 & -0.362 \\
& & 15,27 & 0.8243 & 0.8401 & -0.0158 & -0.358 \\
& & 11,0  & 0.7810 & 0.7935 & -0.0125 & -0.338 \\
& & 21,11 & 0.8887 & 0.8802 & +0.0084 & +0.271 \\

\bottomrule
\end{tabular}
}
\end{table}

We hypothesize hallucinations exhibits as differences in modality-specific attention concentration. To quantify these differences, we employ the Gini coefficient~\cite{gini}, which measures how unevenly attention is distributed across tokens within each modality. We observe a consistent distinction between hallucinated and non-hallucinated responses: hallucinated ones exhibit higher text attention concentration and more diffuse visual attention, as shown in figure~\ref{fig:gini_visual}. Table~\ref{tab:gini_attention} further supports this modality-specific distinction: visual attention is more concentrated in non-hallucinated than in hallucinated responses, as indicated by negative mean Gini differences across most of the selected statistically significant visual heads~\cite{Devil_in_the_middle_layer}. These findings motivate \underline{M}odality-\underline{A}ware \underline{G}ini \underline{I}nspe\underline{C}tion (MAGIC), a hallucination detection framework that decomposes attention by modality and leverages Gini-based concentration measures as an interpretable signal for hallucination detection. Our key contributions are:

\begin{enumerate}
\item We introduce MAGIC, a modality-aware framework that quantifies attention concentration across modalities during answer generation using the Gini coefficient.
\item We show that MAGIC's attention signal can effectively detect hallucinations and benchmark it against four baselines on two medical LMMs across three VQA datasets.
\end{enumerate}

\begin{figure}[t]
\includegraphics[width=\linewidth]{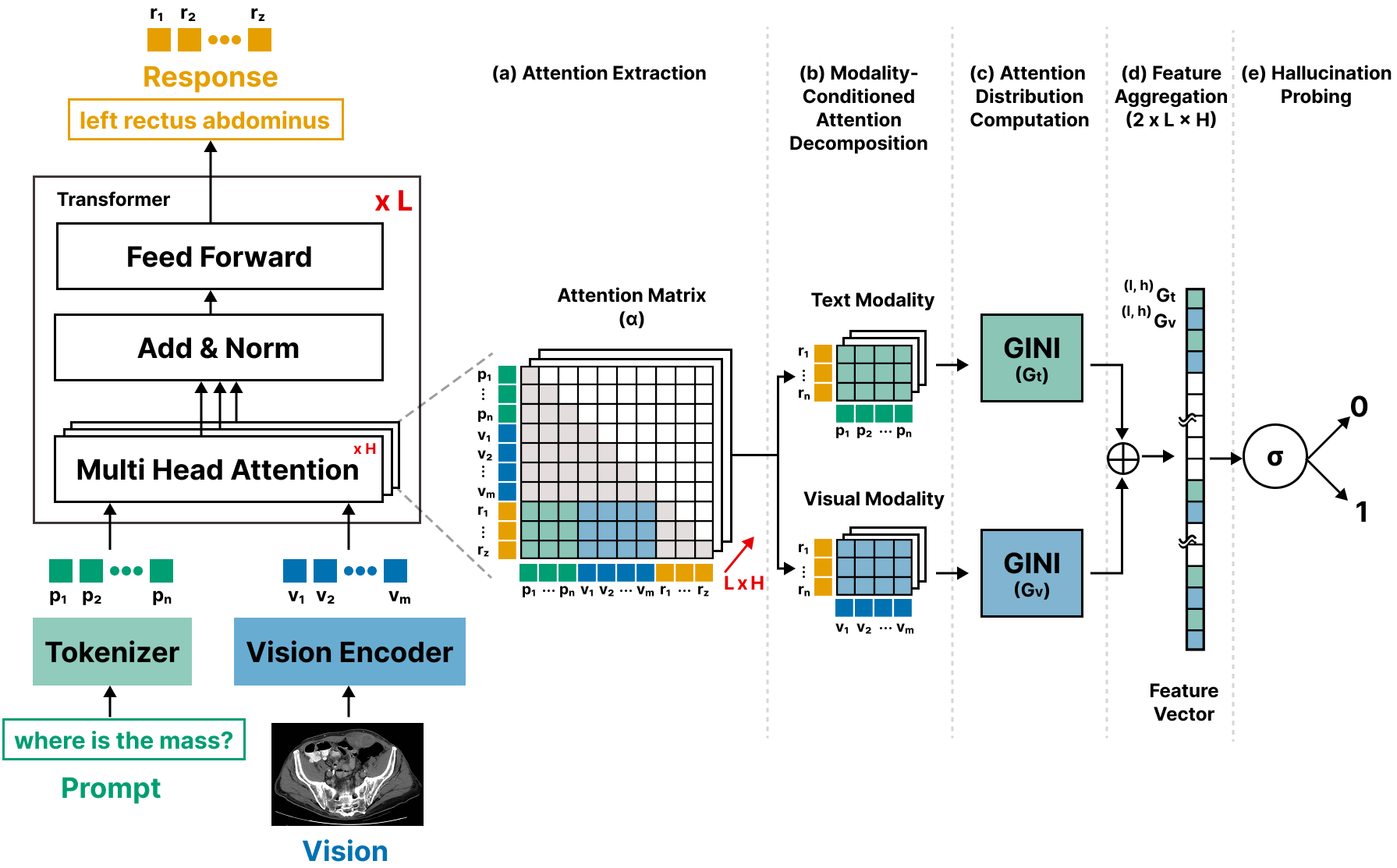}
\caption{
\textbf{MAGIC framework.} Given a multimodal input (image-question pair), (a) attention matrices are extracted across \(L\) layers and \(H\) heads. (b) Each matrix is decomposed into text and visual components, isolating response-token attention to prompt and image tokens. (c) Modality-specific Gini coefficients (\(G_t\), \(G_v\)) quantify attention concentration. (d) The coefficients are aggregated into a \(2 \times L \times H\) feature tensor. (e) A lightweight logistic regression probe \(\sigma\) predicts hallucination.}
\label{fig:methodology}
\end{figure}

\section{Methodology}
\label{sec:method}
\subsection{Preliminary}

Given an image-question pair, the image is encoded into visual tokens $V = \{ v_j \}_{j=1}^{m}$, and the question is tokenized into prompt tokens $P = \{ p_k \}_{k=1}^{n}$. The modality connector projects the visual tokens into the language model's embedding space. The projected tokens are then concatenated with the prompt tokens and fed into the language model, which generates response tokens autoregressively as in Equation~\ref{equation:autoregressive}.

\begin{equation}
    \label{equation:autoregressive}
    r_i = \arg\max_{w \,\in\, \mathcal{W}} \; p_\theta(w \mid P, V,r_{<i})
\end{equation}
\noindent where $r_i$ denotes the $i$-th response token for $i \in \{1, \dots, z\}$ and $\mathcal{W}$ denotes the vocabulary.

The language model is composed of $L$ transformer layers with $H$ attention heads each. For a given layer $l$ and head $h$, the attention is computed as,
\begin{equation}
    \label{equation:attention}
    \alpha^{(l,h)} = \text{Softmax}\left(\frac{\mathbf{Q}\mathbf{K}^T}{\sqrt{d_k}}\right)
\end{equation}
\noindent where $\mathbf{Q}, \mathbf{K} \in \mathbb{R}^{s \times d_k}$ are the query and key matrices, $s=n+m+z$ denotes the total sequence length and $d_k$ denotes their dimensionality. Since the attention matrix at the final generation step aggregates information from both modalities, we extract it to analyze cross-modal interaction and modality-specific attention matrix.

\subsection{Modality-Specific Attention Decomposition}

We extract modality-specific sub-matrices from the attention weights to quantify the influence of textual and visual tokens on each generated response token. We analyze the attention weight matrix \(\alpha^{(l,h)} \in \mathbb{R}^{s \times s}\) at layer \(l\) and head \(h\). Considering the concatenated token sequence \([P, V, R]\), the text-modality attention sub-matrix and visual-modality sub-matrix is then defined as
\begin{equation}
\begin{aligned}
\text{Text}^{(l,h)} &= \bigl[ \alpha^{(l,h)}_{i,j} \bigr]_{i \in R,\, j \in P} \in \mathbb{R}^{z \times n}, \\
\text{Visual}^{(l,h)} &= \bigl[ \alpha^{(l,h)}_{i,j} \bigr]_{i \in R,\, j \in V} \in \mathbb{R}^{z \times m}.
\end{aligned}
\end{equation}

Here, \(\text{Text}^{(l,h)}_{i,j}\) represents the attention weight assigned to the \(j\)-th prompt token when generating the \(i\)-th response token, while \(\text{Visual}^{(l,h)}_{i,j}\) represents the attention weight assigned to the \(j\)-th visual token for the i-th response token.

\subsection{Gini Coefficient}
To quantify attention concentration, we compute the Gini coefficient, a scalar measure of distributional inequality. For a sorted distribution $X=\{x_i\}_{i=1}^N$,  the Gini coefficient is computed as 
\begin{equation}
    \label{equation:gini_efficient}
    \text{Gini($X$)} = \frac{2\sum_{i=1}^Ni\cdot x_i}{N\sum_{i=1}^N
    x_i}-\frac{N+1}{N}
\end{equation}
Higher values indicate more concentrated attention, while lower values indicate a more uniform distribution.
 
 For each layer $l$ and head $h$, we compute modality-specific Gini coefficients over the flattened visual and textual attention matrices:
\begin{equation}
G_m^{(l,h)} = \text{Gini}\big(\text{flatten}(A_m^{(l,h)})\big), \quad m\in{v,t},
\end{equation}
where $A_v$ and $A_t$ denote visual and textual attention, respectively. The resulting coefficients are concatenated across all layers and heads to form the feature vector:
\begin{equation}
F = \big[G_v^{(1,1)},G_t^{(1,1)},\ldots,G_v^{(L,H)},G_t^{(L,H)}\big],
\end{equation}
as illustrated in figure~\ref{fig:methodology}.

\subsection{Hallucination Probe}
A lightweight logistic regression probe is trained on the Gini-based features to predict hallucination:

\begin{equation}
\hat{y} = \sigma(\mathbf{w}^\top F + b), \quad y \in \{0,1\}.
\end{equation}

\noindent Here, $y=1$ denotes a hallucinated response, and $y=0$ a faithful response.  

\section{Experiments and Results}
\label{sec:experiments}
\subsection{Setup}
\noindent \textbf{Datasets and LMMs:}
We evaluate our framework on three medical VQA benchmarks: VQA-RAD~\cite{radvqa}, PathVQA~\cite{pathvqa}, and Kvasir-VQA-X1~\cite{kvasirvqax1}, using LLaVA-Med-7B~\cite{LLava_med} and MedGemma-4B~\cite{Med_gemma}. We randomly sample 5,000 training instances from PathVQA and Kvasir-VQA-X1 and use the full VQA-RAD training split (1,793 samples), while evaluating on the complete official test splits. For attention extraction, we use all layers and heads of LLaVA-Med and only global attention layers of MedGemma to ensure full-sequence contextual access. All models use greedy decoding (temperature = 0), and hallucination labels are derived from GREEN~\cite{Green_model} scores using a threshold of less than 1, following prior validation~\cite{kvasirvqax1}. Experiments are conducted on NVIDIA A100 GPUs, with logistic regression implemented in scikit-learn. We compare against AttenEigen~\cite{spectral_analysis} and LaplacianEigen~\cite{spectral_analysis}, as well as AttenVisual and AttenFull adapted from Lookback Lens~\cite{look_back_lens}. Performance is evaluated using AUROC and AUPR.

\subsection{Results}
\noindent \textbf{Comparison Against Baselines:}
\begin{table}[t]
\caption{Hallucination detection performance. 
}
\label{tab:main_results}
\setlength{\tabcolsep}{3pt}
\fontsize{8}{9}\selectfont
\resizebox{\linewidth}{!}{%
\begin{tabular}{lcccccc}
\toprule
\multirow{2}{*}{Method} 
& \multicolumn{3}{c}{LLaVA-Med-7B} 
& \multicolumn{3}{c}{MedGemma-4B} \\
\cmidrule(lr){2-4} \cmidrule(lr){5-7}
& VQARAD & PathVQA & KvasirVQA-X1 
& VQARAD & PathVQA & KvasirVQA-X1 \\
\midrule
\multicolumn{7}{c}{\textbf{AUROC}} \\
\midrule
AttenEigen\cite{spectral_analysis}      & 0.6717 & 0.8503 & 0.8891 & 0.6306 & 0.7908 & 0.7143 \\
LaplacianEigen\cite{spectral_analysis}  & 0.7033 & 0.8433 & 0.8905 & 0.6476 & 0.8128 & 0.7730 \\
AttenFull\cite{look_back_lens}           & 0.7080 & \textbf{0.8695} & 0.8183 & 0.6714 & 0.8224 & 0.8184 \\
AttenVisual\cite{look_back_lens}         & 0.7051 & 0.8554 & 0.8305 & 0.6920 & 0.8261 & 0.8305 \\
\textbf{MAGIC(Ours)}                             & \textbf{0.7229} & 0.8574 & \textbf{0.8944} & \textbf{0.6941} & \textbf{0.8344} & \textbf{0.8457} \\
\midrule
\multicolumn{7}{c}{\textbf{AUPR}} \\
\midrule
AttenEigen\cite{spectral_analysis}      & 0.7039 & 0.9292 & 0.9677 & 0.5936 & 0.8700 & 0.9244 \\
LaplacianEigen\cite{spectral_analysis}  & 0.7286 & 0.9297 & 0.9675 & 0.6227 & 0.8755 & 0.9353 \\
AttenFull\cite{look_back_lens}           & 0.7319 & \textbf{0.9408} & 0.9518 & 0.6662 & 0.8900 & 0.9518 \\
AttenVisual\cite{look_back_lens}         & 0.7351 & 0.9322 & 0.9519 & \textbf{0.6884} & 0.8940 & 0.9519 \\
\textbf{MAGIC(Ours)}                             & \textbf{0.7383} & 0.9352 & \textbf{0.9711} & 0.6823 & \textbf{0.8996} & \textbf{0.9565} \\
\bottomrule
\end{tabular}%
}
\end{table}
Table~\ref{tab:main_results} reports hallucination detection performance on LLaVA-Med-7B and MedGemma-4B across three medical VQA datasets. Our method achieves the highest AUROC on KvasirVQA-X1 for both models (0.8944 for LLaVA-Med-7B, 0.8457 for MedGemma-4B) and the highest AUPR on the same dataset  (0.9711 and 0.9565, respectively). On VQA-RAD, it reaches the top AUROC for LLaVA-Med-7B (0.7229) and MedGemma-4B (0.6941), while on PathVQA, the AUROC is 0.8574 for LLaVA-Med-7B and 0.8344 for MedGemma-4B, close to the best-performing baselines. These results show that our method consistently performs well across two model architectures and three imaging domains.

\subsection{Ablation Study}
\begin{table}[t]
\caption{Significance of attention decomposition.}
\label{ablation_decomposition}
\setlength{\tabcolsep}{3pt}
\fontsize{8}{9}\selectfont
\resizebox{\linewidth}{!}{
\begin{tabular}{lcccccc}
\toprule
\multirow{2}{*}{Method} 
& \multicolumn{3}{c}{LLaVA-Med-7B} 
& \multicolumn{3}{c}{MedGemma-4B} \\
\cmidrule(lr){2-4} \cmidrule(lr){5-7}
& VQARAD & PathVQA & KvasirVQA-X1 
& VQARAD & PathVQA & KvasirVQA-X1 \\
\cmidrule(lr){2-7}
& \multicolumn{6}{c}{AUROC} \\
\midrule
Full Attention      & 0.5355 & 0.5827 & 0.6372 & 0.5770 & 0.7231 & 0.6269 \\
Text Attention      & 0.6959 & 0.8118 & 0.8634 & 0.6863 & 0.8210 & 0.8098 \\
Visual Attention      & 0.7129 & 0.8520 & 0.8899 & 0.6729 & 0.8257 & 0.8377 \\
\textbf{MAGIC(Ours)}      & \textbf{0.7229} & \textbf{0.8574} & \textbf{0.8944} & \textbf{0.6941} & \textbf{0.8344} & \textbf{0.8457} \\
\midrule
& \multicolumn{6}{c}{AUPR} \\
\midrule
Full Attention      & 0.5563 & 0.7464 & 0.8656 & 0.5458 & 0.8482 & 0.8925 \\
Text Attention      & 0.7265 & 0.9173 & 0.9621 & 0.6728 & 0.8818 & 0.9400 \\
Visual Attention      & 0.7309 & 0.9329 & 0.9694 & 0.6635 & 0.8950 & \textbf{0.9568} \\
\textbf{MAGIC(Ours)}      & \textbf{0.7383} & \textbf{0.9352} & \textbf{0.9711} & \textbf{0.6823} & \textbf{0.8996} & 0.9565 \\
\bottomrule
\end{tabular}
}
\end{table}
Table~\ref{ablation_decomposition} reports the significance of attention decomposition on hallucination detection performance across both models and all three benchmark datasets. Using the full, undifferentiated attention matrix yields the weakest results, suggesting that the full attention signals introduce noise that obscures meaningful attention patterns. Decomposing attention by modality yields substantial improvements: visual and textual attention each independently outperform the full attention baseline, and combining both achieves the best AUROC in all six evaluated settings and the best AUPR in five of the six, suggesting that the modalities capture complementary signals that can be exploited jointly for hallucination detection.

\section{Conclusion}
\label{sec:conclusion}
We introduced MAGIC, a modality-aware attention-based framework for hallucination detection in medical LMMs. By decomposing attention into visual and textual components and quantifying their concentration using the Gini coefficient, MAGIC identifies modality imbalance patterns associated with hallucinations. Evaluations across multiple medical VQA benchmarks and two LMM architectures demonstrate consistent detection performance, either outperforming or matching prior attention-based approaches. Moreover, MAGIC provides interpretable insights into modality-specific attention, helping assess whether responses are visually grounded or dominated by textual information. Future work could extend MAGIC to hallucination mitigation through intervention-aware decoding strategies.

\section{Acknowledgments}
The authors acknowledge the use of AI assistants for grammatical correction, improving manuscript coherence, and coding assistance.


\end{document}